\documentclass[sigconf]{acmart}

\usepackage{graphicx}
\usepackage{lipsum}
\usepackage{indentfirst}
\usepackage{amsmath}
\usepackage{cases}
\usepackage{booktabs}
\usepackage{multirow} 
\usepackage{graphicx}
\usepackage{times}  
\usepackage{algorithm}
\usepackage{algorithmic}
\usepackage{color}
\usepackage{ulem}
\usepackage{pifont}
\usepackage{arydshln} 
\usepackage{float} 
\usepackage{bm}
\usepackage{caption}

\AtBeginDocument{%
  }

\acmISBN{978-1-4503-XXXX-X/2018/06}

\begin{document}

\title{Stable-Makeup: When Real-World Makeup Transfer Meets Diffusion Model}


\author{Yuxuan Zhang}
\email{zyx153@sjtu.edu.cn}
\affiliation{%
  \institution{Shanghai Jiao Tong University}
  \city{Shanghai}
  \country{China}}

\author{Yirui Yuan}
\email{yuanyr2023@shanghaitech.edu.cn}
\affiliation{%
  \institution{ShanghaiTech University}
  \city{Shanghai}
  \country{China}}

\author{Yiren Song}
\email{yiren@nus.edu.sg}
\affiliation{%
  \institution{National University of Singapore}
  \country{Singapore}
}

\author{Jiaming Liu}
\authornote{Corresponding author.}
\email{jmliu1217@gmail.com}
\affiliation{%
  \institution{Tiamat}
  \city{Shanghai}
  \country{China}}
  
\renewcommand{\shortauthors}{Trovato et al.}



\begin{abstract}
\label{abs}
  Current makeup transfer methods are limited to simple makeup styles, making them difficult to apply in real-world scenarios. In this paper, we introduce Stable-Makeup, a novel diffusion-based makeup transfer method capable of robustly transferring a wide range of real-world makeup, onto user-provided faces. Stable-Makeup is based on a pre-trained diffusion model and utilizes a Detail-Preserving (D-P) makeup encoder to encode makeup details. It also employs content and structural control modules to preserve the content and structural information of the source image. With the aid of our newly added makeup cross-attention layers in U-Net, we can accurately transfer the detailed makeup to the corresponding position in the source image. After content-structure decoupling training, Stable-Makeup can maintain the content and the facial structure of the source image. Moreover, our method has demonstrated strong robustness and generalizability, making it applicable to various tasks such as cross-domain makeup transfer, makeup-guided text-to-image generation, and so on. Extensive experiments have demonstrated that our approach delivers state-of-the-art results among existing makeup transfer methods and exhibits a highly promising with broad potential applications in various related fields. Code released: \url{https://github.com/Xiaojiu-z/Stable-Makeup}
  
\end{abstract}

\begin{teaserfigure}
\centering
  \includegraphics[width=1\textwidth]{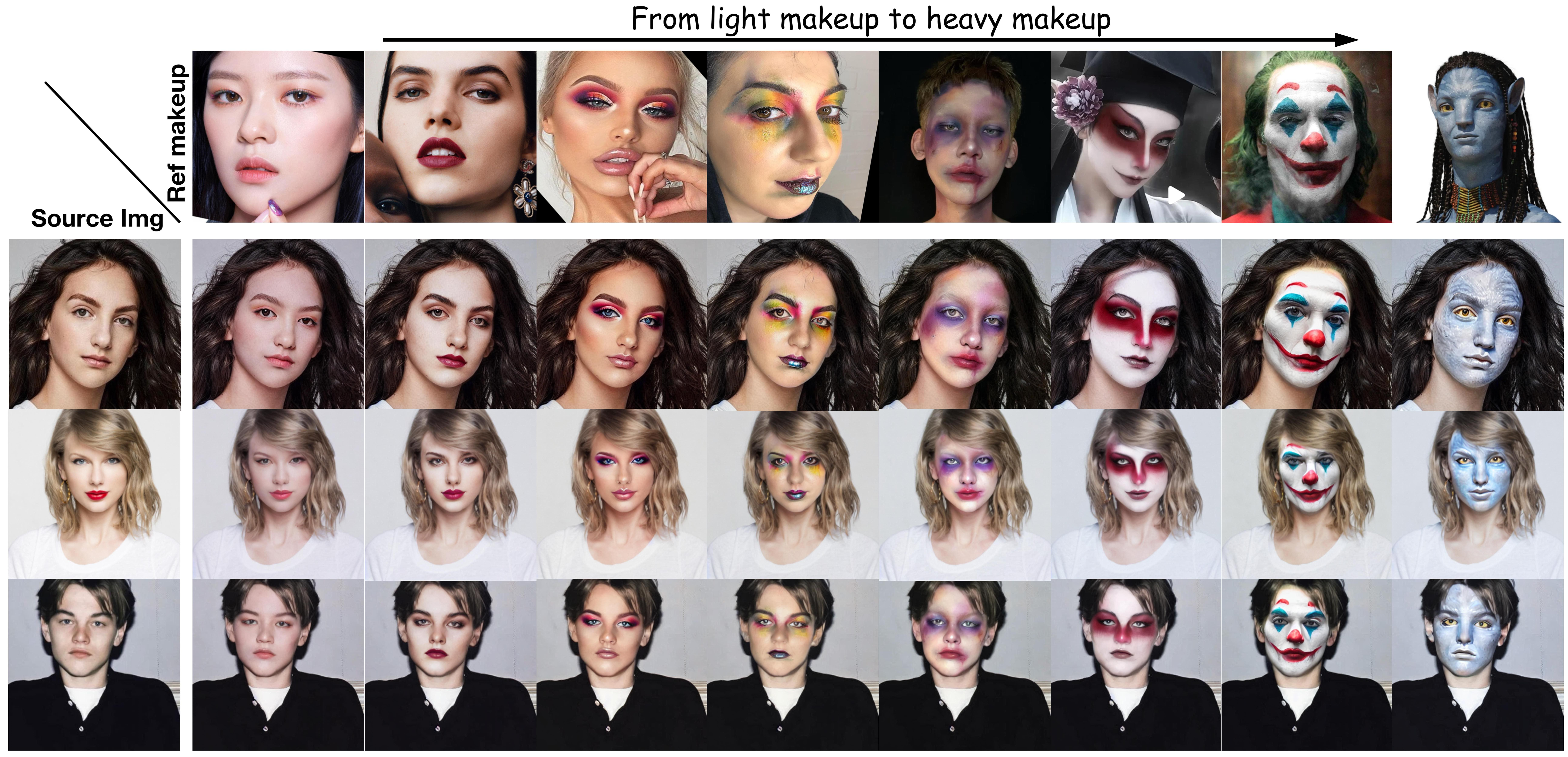}
  \vspace{-0.5cm}
  \caption{Our proposed framework, \textit{Stable-Makeup}, is a novel diffusion-based method for makeup transfer that can robustly transfer a diverse range of real-world makeup styles, from light to extremely heavy makeup.}
  \Description{}
  \label{fig:teaser}
\end{teaserfigure}

\maketitle
\section{Introduction}
\label{sec:intro}
As a significant computer vision task, makeup transfer has a wide range of applications, such as in the beauty industry or in virtual try-on systems, and enables the enhancement, modification, and transformation of facial features, achieving the desired effects of beautification, embellishment, and deformation. However, despite its conceptual straightforwardness, makeup transfer poses significant challenges when aiming for a seamless and authentic transformation across a wide range of makeup intensities and styles.

From the perspective of the existing technical route, current makeup transfer techniques~\cite{li2018beautygan, jiang2020psgan, deng2021spatially, nguyen2021lipstick, xiang2022ramgan, gu2019ladn, liu2021psgan++, wan2022facial, yan2023beautyrec, sun2023ssat, kips2020gan, yang2022elegant} rely heavily on Generative Adversarial Network (GAN) based approaches which have demonstrated potential in digital cosmetic applications. However, these approaches fall short when confronted with the diversity of real-world makeup styles, especially when translating high-detailed and creative cosmetics, such as those found in cosplay or movie character imitations, onto real faces. This limitation not only restricts their applicability but also hinders their effectiveness in accurately capturing the essence of personalized and intricate makeup designs. Recognizing this gap, our research introduces Stable-Makeup, a novel approach leveraging diffusion-based methodologies to transcend the boundaries of existing makeup transfer methods.

From a data perspective, existing makeup datasets lack diversity and cannot accommodate real-world makeup transfer. To address this data shortage, we propose an automatic data construction pipeline that employs a large language model and generative model to edit real human face images and create paired before-and-after makeup images. Ultimately, our training data comprises 20k image pairs, covering a wide range of makeup styles from light to heavy makeup. This automatic data construction pipeline has the potential to facilitate research in the makeup transfer field and enable the development of more robust and accurate models.

Our Stable-Makeup, which is built on a pre-trained diffusion model, comprises three key components: the Detail-Preserving Makeup Encoder, Makeup Cross-attention Layers, and the Content and Structural Control Modules. To preserve makeup details, we employ a multi-layer strategy in our D-P makeup encoder to encode the makeup reference image into multi-scale detail makeup embeddings. The content control module is designed to maintain pixel-level content consistency with the source image. The structural control module is utilized to introduce facial structure, improving the consistency between the generated image and the facial structure of the source image. To achieve semantic alignment between the intermediate features of the Unet-encoded source image and the detail makeup embeddings, we extended the U-Net architecture by incorporating a makeup branch composed of cross-attention layers. Through our proposed content and structural decoupling training strategy, we can further maintain the facial structure of the source image. 
As exhibited in Fig.~\ref{fig:teaser}, By integrating a meticulously curated dataset alongside a comprehensive pipeline from data acquisition to model training, we lay the foundation for a robust framework capable of handling the nuanced dynamics of makeup transfer with unprecedented precision and adaptability.

In summary, our contributions are: 
\begin{itemize}
\item \textit{Stable-Makeup} is the first diffusion-based makeup transfer method to the best of our knowledge. Our experimental results demonstrate state-of-the-art performance over existing makeup transfer methods. Moreover, Stable-Makeup demonstrates high robustness and generalizability, making it applicable to various tasks.
\item To ensure the accuracy and consistency of makeup transfer, we propose a Detail-Preserving makeup encoder in conjunction with makeup cross-attention layers that align the semantic correspondence between detailed makeup features and human face intermediate features in U-Net. 
\item We designed an automatic pipeline to create a high-quality, diverse paired dataset of makeup before and after images. After content-structure decoupling training on our training data, we can further maintain the content and structure of the source image.
\end{itemize}

\section{Related Works}
\label{related}
\subsection{Facial Makeup Transfer}
Traditional makeup transfer methods employ image processing techniques such as facial landmark extraction and detection\cite{xu2013automatic, tong2007example}. Meanwhile, contemporary deep learning-based makeup transfer methods demonstrate high robustness and generalizability to diverse makeup styles, Generative Adversarial Networks (GANs)\cite{zhu2017unpaired, chang2018pairedcyclegan, li2018beautygan, jiang2020psgan, deng2021spatially, nguyen2021lipstick, xiang2022ramgan, gu2019ladn, liu2021psgan++, wan2022facial, yan2023beautyrec, sun2023ssat, kips2020gan, hu2022protecting, yang2022elegant} as an example, have been extensively applied in facial makeup synthesis tasks. 

In the domain of image transfer, Beauty-GAN\cite{li2018beautygan} is driven by pixel-level Histogram Matching and utilizes an array of loss functions to train the primary network. However, its effectiveness is limited to frontal facial images, posing a challenge when transferring makeup between images with notable facial pose variations. PSGAN\cite{jiang2020psgan} addresses the challenge of transferring makeup between images with significant facial expression variations and focuses on specific facial regions for transfer. To expand the scope of makeup transfer tasks beyond mere color transfer in specific facial areas, CPM\cite{nguyen2021lipstick} introduces pattern addition into the makeup transfer process. Furthermore, SCGAN\cite{deng2021spatially} leverages a Part-specific Style Encoder to decouple makeup styles on a component-wise basis.  Additionally, RamGAN\cite{xiang2022ramgan} focuses on component consistency by integrating a region-attentive morphing module. Although these GAN-based methods have achieved significant success, they exhibit limitations in transferring complex makeup styles in real-world scenarios. Moreover, the majority of these approaches necessitate the alignment of facial regions, further constraining their practical applicability.

The inadequacies of GAN-based models in addressing makeup transfer problems under real-world makeup settings have necessitated the development of new techniques. The primary objective of this paper is to overcome these limitations by utilizing diffusion-based methods to achieve robust and high-quality output for extreme style makeup.

\subsection{Diffusion Models}
Recently, diffusion models have ushered in significant advancements in multi-modal image generation tasks, encompassing text-to-image generation\cite{dalle2, sdxl, Imagen, rombach2022high, IF}, image editing\cite{attendandexcite, layerdiffusion, selfguidance, masactrl,editevery,dragondiffusion,iedit,sine,tsaban2023ledits}, controllable generation\cite{controlnet,t2i,unicontrol,directed} and subject-driven generation\cite{zhang2024ssr, lora, domainagnostic,taming,TI, hyperdreambooth, DB}. Methods based on the diffusion paradigm, along with their subsequent adaptations, have demonstrated state-of-the-art results in various image generation tasks.

As for controllable generation, ControlNet\cite{zhang2023adding} employs a bypass structure featuring a trainable copy network as the input mechanism for control conditions, effectively integrating a variety of control conditions such as Canny Edge, Depth, and Human Pose, etc. As for subject generation, DreamBooth\cite{DB} Dreambooth trains a separate Unet for each subject, resulting in excellent consistency of character. Meanwhile, diffusion-based image editing methods have also demonstrated remarkable progress. For example, Ledits\cite{tsaban2023ledits} introduces a training-free image manipulation method, utilizing an inversion approach for processing.

Due to the remarkable advancements of diffusion methods in generating high-quality images, benefiting from supervised training on large-scale datasets, they have become the mainstream method in image generation. Therefore, in this paper, we have leveraged the advantages provided by the diffusion model to the fullest extent in the task of makeup transfer. Our approach overcomes the limitations of GAN-based models and achieves robust and high-quality output, particularly for extreme style makeup.
\section{Methodology}
\label{method}
In this part, we introduce the detailed design of our Stable-Makeup. Then, the specifics of our training and inference process are in a detailed description. Preliminaries about diffusion models~\cite{rombach2022high}, Controlnet~\cite{zhang2023adding} and CLIP~\cite{clip} can be found in supplementary.

\subsection{Stable-Makeup}
\label{stablemakeup}
\subsubsection{Overview.}

\begin{figure*}[t]
\centering
\vspace{-0.3cm}
\includegraphics[width=0.9\linewidth]{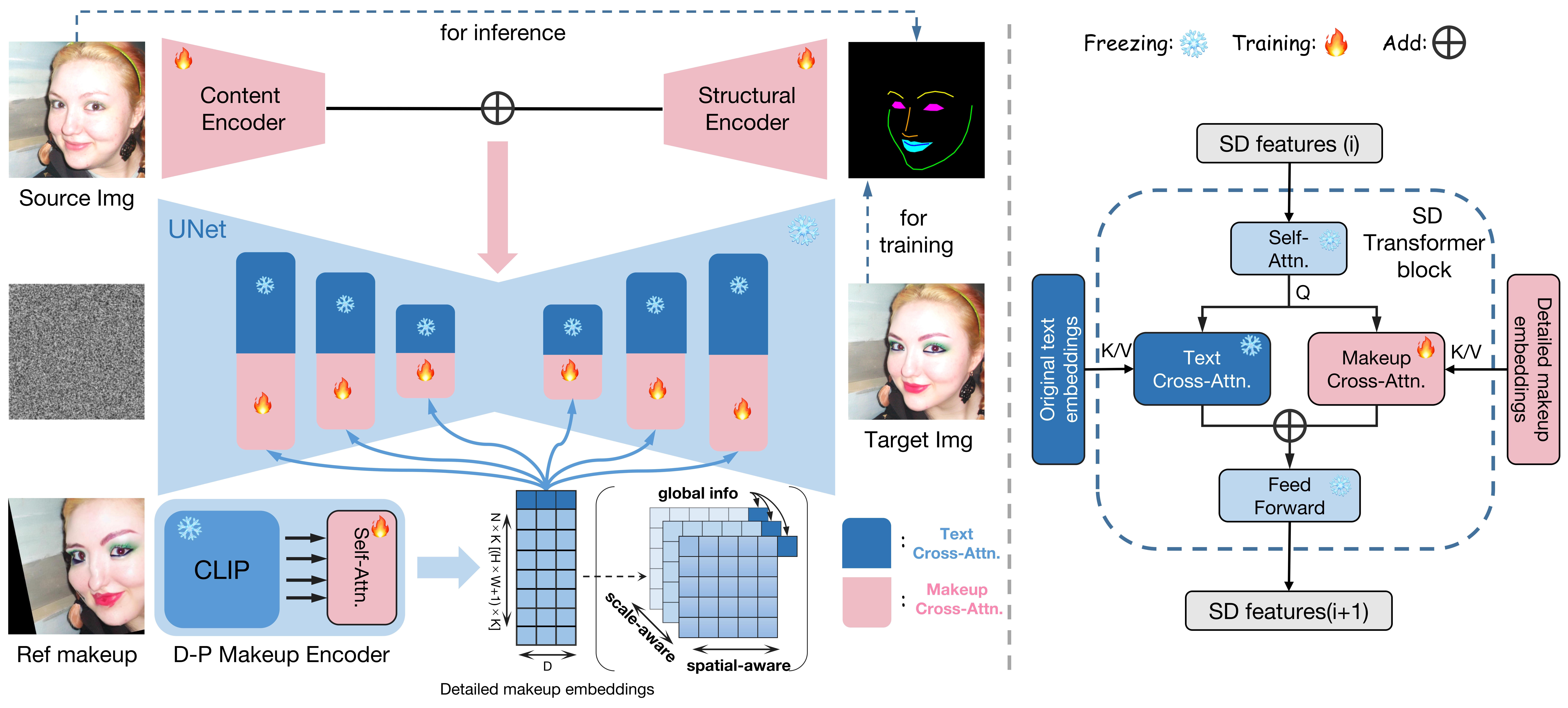}
\vspace{-0.3cm}
  \caption{Overall schematics of our method. Given a source image $\mathit{I_s}$, a reference makeup image $\mathit{I_m}$ and an obtained facial structure control image $\mathit{I_c}$, Stable-Makeup utilizes the D-P makeup encoder to encode $\mathit{I_m}$. Content and structural encoders are used to encode $\mathit{I_s}$ and $\mathit{I_c}$ respectively. With the aid of the makeup cross-attention layers, Stable-Makeup aligns the facial regions of $\mathit{I_s}$ and $\mathit{I_m}$, enabling successful transfers of the intricate makeup details. After content-structure decoupling training, Stable-Makeup further maintains content and structure of $\mathit{I_s}$.}
    \label{fig:method}
    \vspace{-0.4cm}
\end{figure*}

The overall methodology is illustrated in Fig.~\ref{fig:method}. Formally, the target of Stable-Makeup is to transfer the reference makeup from a given image $\mathit{I_m}$ onto a user-provided source image $\mathit{I_s}$, resulting in a target image $\mathit{I_t}$ that exhibits the desired makeup style. Therefore, Stable-Makeup necessitates the simultaneous transfer of fine-grained makeup information and the preservation of structural information. To achieve this, Stable-Makeup first utilizes a Detail-Preserving Makeup Encoder to extract the multi-scale features of the reference makeup as detailed makeup embeddings $\mathit{E_m}$. Next, content encoder and structural encoder are employed to encode the source image as content features $\mathit{F_c}$ and the facial structure control image $\mathit{I_c}$ (obtained from an off-the-shelf face parsing method~\cite{spiga}.) as structure features $\mathit{F_s}$, respectively. These two features are then incorporated together and fed into the U-Net. The multi-scale detailed makeup embeddings $\mathit{E_m}$ along with the content features $\mathit{F_c}$ and structure features $\mathit{F_s}$ are integrated into the diffusion U-Net as conditions to guide the image generation process. During target image $\mathit{I_t}$ generation, Stable-Makeup uses newly added cross-attention layers to implicitly align the multi-scale detailed makeup embeddings $\mathit{E_m}$ with the intermediate feature maps of facial region in source image at different layers in U-Net. The alignment ensures that the generated target image contains both the detailed information of the reference makeup and maintains the content and structure of the source image $\mathit{I_s}$. With our Content and Structure Decoupling Training, Stable-Makeup can further maintain the content and facial structure of the source image. 

\subsubsection{Detail-Preserving Makeup Encoder.}
\label{sec:detail}
Makeup images often involve intricate details such as eyelashes, eyebrows, and fine lines. These details are often subtle and precise, making it challenging to maintain their quality during makeup transfer. To address this issue, we have carefully designed a Detail-Preserving Makeup Encoder to capture the details of the makeup.

Inspired by~\cite{ye2023ip, zhang2024ssr}, we employ a pre-trained CLIP~\cite{clip} visual backbone to extract makeup representations from reference images $\mathit{I_m}$. As mentioned in~\cite{zhang2024ssr}, extracting visual embeddings from the last CLIP layer cannot preserve fine details of the reference image. To address this issue, we extract features across various layers from the CLIP visual backbone and concatenate them along the feature dimension. To improve the performance of makeup transfer, we utilize a self-attention layer to process the multi-layer features encoded by CLIP, which helps the model better capture both local and global features of the makeup. This results in the generation of detailed makeup embeddings that are used for subsequent makeup cross-attention layers, fusing with the target image features to generate the final transfer results. Formally, the Makeup Encoder processes a reference makeup image $\mathit{I_m}$ to produce multi-scale detailed makeup embeddings $\mathit{E_m} =  \operatorname{concat}_{k=0}^{K} \left(\mathit{E^k, dim=1}\right)$, where $\mathit{E^k}$ is image embeddings at the layer of $k$ in CLIP visual backbone and $K$ refers to the number of layers we extract. In our experimental setup, we fixed the value of K to 12.

Notably, some previous works\cite{ye2023ip,elite,e4t} utilize linear layer mapping of CLIP features. This design cannot preserve the spatial information between input features. Different from them, we utilize a self-attention layer to map the multi-layer CLIP embeddings (256-dimensional patch features and 1-dimensional CLS features) of the reference makeup image and concatenate on feature dimension. This approach effectively preserves the spatial, multi-scale, and global information of the makeup image, thereby facilitating improved alignment of intermediate-related features within the U-Net architecture. Consequently, it yields more precise preservation of fine-grained details and enhanced transfer outcomes. Combining both spatial and scale feature maps, our D-P makeup encoder yields the most complete information for makeup. Furthermore, we provide additional comparative analysis in supplementary to demonstrate the superiority of our method.

\subsubsection{Makeup Cross-attention Layers.}
The goal of the makeup cross-attention layers is to integrate multi-scale detailed makeup embeddings into the U-Net and align them appropriately with the intermediate feature maps ofthe  facial region in the source image, enabling the model to accurately transfer the makeup onto the facial region in the source image. Specifically, as shown in the figure~\ref{fig:method}, in each transformer block of the U-Net, we retain the original text cross-attention and add Makeup Cross-attention Layers. The multi-scale detailed makeup embeddings $\mathit{E_m}$ are input to the makeup cross-attention layers and serve as the K(key) and V(value) features. The makeup cross-attention layers and the text cross-attention layers in the original U-Net share the Q feature. With the help of the cross-attention mechanism, the intermediate facial features of the source image and makeup image are well-aligned in the stable diffusion U-Net. Finally, we simply add the output of makeup cross-attention to the output of text cross-attention. 

Our design of adding the output of the cross-attention between the makeup and the original text preserves the original text-guided generation capability. In the context of the classic makeup transfer task, the influence of text is not considered, only the makeup cross-attention is employed during both training and inference. However, our design opens up a new direction, namely makeup-guided text-to-image generation, as illustrated in Figure~\ref{fig:app}. In this task, the reference makeup image is employed as a guiding condition in conjunction with the original text prompt to influence the pre-trained U-Net architecture, and the makeup consistency is maintained during the text-to-image generation process.

\subsubsection{Content and Structural Control Modules.}
To maintain the structural consistency of the source image during makeup transfer, we utilize Content and Structural Control Modules, which consist of two adapted ControlNets serving as encoders. The target of the content encoder is to maintain content consistency and consistency in non-facial areas of the original image, and it uses ControlNet to encode the source image. To control facial structures during the makeup transfer process, we added an extra structural encoder. In order to increase spatial constraints and introduce more facial information, such as facial shape or mouth closure, we utilized dense lines of varying colors that were drawn based on the different locations of facial key points. These lines served as conditions for the makeup transfer process, enabling us to achieve great control over the resulting facial structures. We use the text embedding of empty text as the cross-attention input for the two ControlNet encoders. Finally, we directly add the outputs of the content encoder and structural encoder to the Stable Diffusion model.

\subsection{Model Training and Inference}
\label{training}

\subsubsection{Training data collection.}
\begin{figure}
    \centering
    \includegraphics[width=0.8\linewidth]{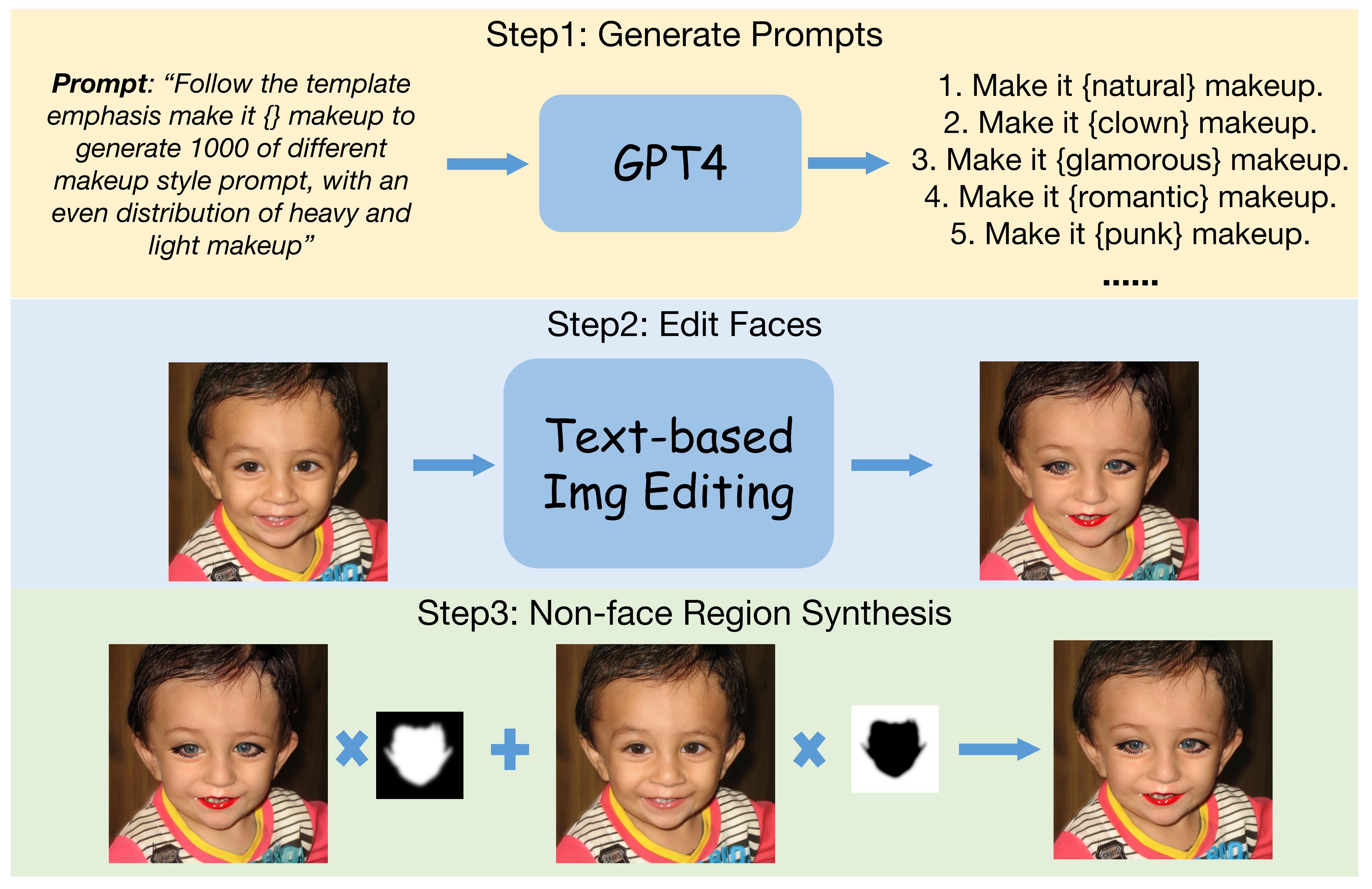}
    \vspace{-0.4cm}
    \caption{The procedure for training data generation.}
     \vspace{-0.5cm}
    \label{fig:data}
\end{figure}

As depicted in Fig.~\ref{fig:data}, our pipeline to create a large-scale makeup pairing dataset by leveraging the capabilities of LLM and pre-trained diffusion involves three main steps: Firstly, we utilize GPT4 to generate 1000 prompts using the template ``Make it \{\} makeup" to cover a wide range of makeup styles. After filtering out duplicate words, we obtain 300 different prompts. Secondly, we sample 20,000 faces from the FFHQ\cite{ffhq} dataset as non-makeup images and apply makeup to these images using existing text-based real image editing algorithm~\cite{tsaban2023ledits}. For each face image, we randomly select a makeup style from the 300 prompts obtained in the first step, resulting in 20,000 face images with consistent identities and makeup styles. Finally, to prevent the unwanted effects of editing non-face regions during the editing process, we use a face segmentation model to separate the face region from the background and other non-face regions. We composite the non-face regions of the original face images onto the edited face images to obtain the final makeup pairing dataset.

Finally, our training data includes 20,000 image pairs covering a wide range of makeup styles from light to heavy makeup, which represents a significant improvement over previous datasets in terms of diversity and quantity. In our supplementary materials, we also present additional visual results for the alternative methods in Step 2.

\begin{figure*}[!h]
\centering
    \vspace{-0.3cm}
  \includegraphics[width=0.8\linewidth]{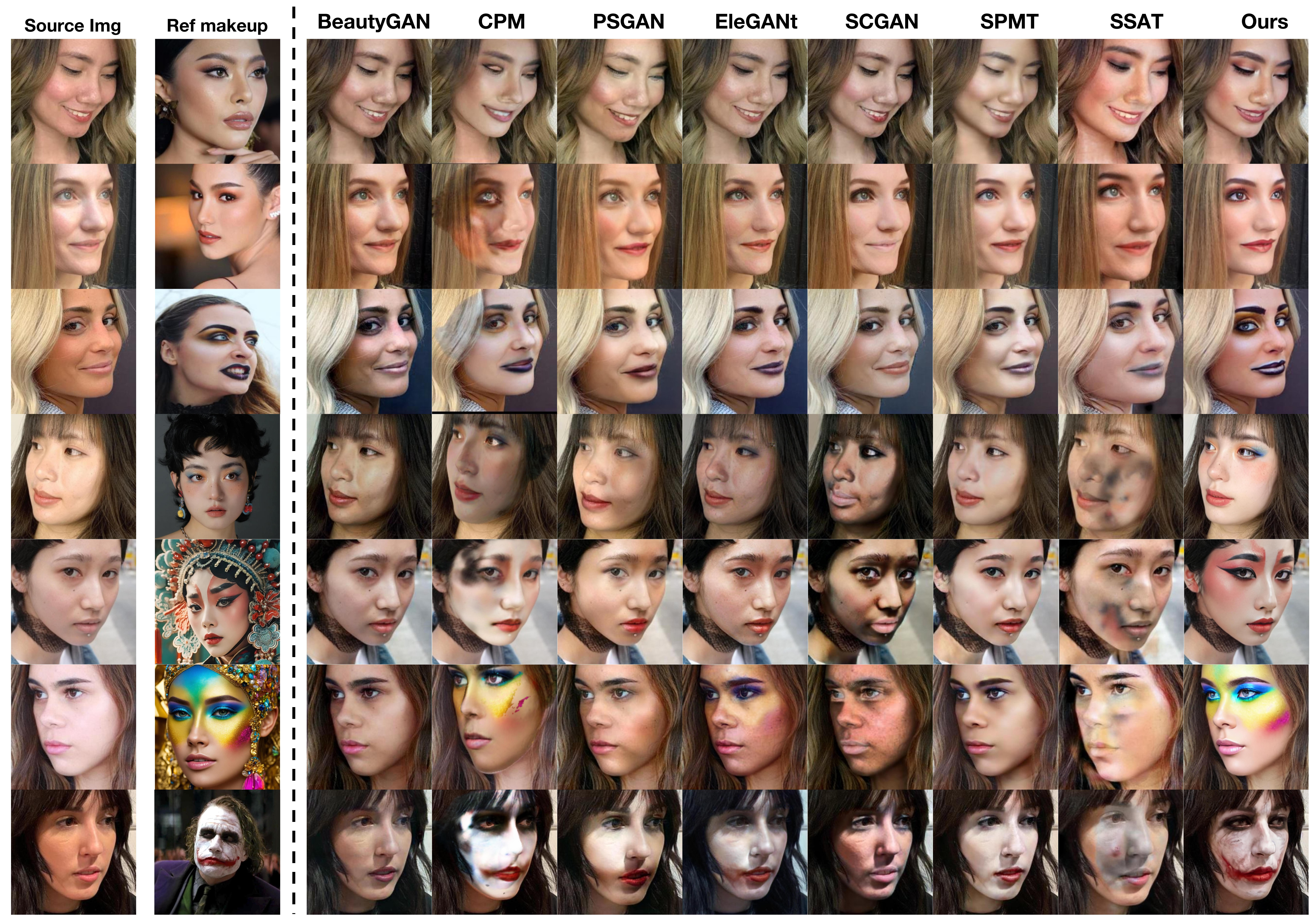}
  \vspace{-0.3cm}
  \caption{Qualitative comparison of different methods. Our results outperform other methods in terms of makeup detail transfer, ranging from light makeup to personalized heavy makeup.}
  \vspace{-0.4cm}
\label{fig:comparison}
\end{figure*}

\subsubsection{Content-Structure Decoupling Training.}
The current diffusion-based real image editing faces a conflict between maintaining the original layout and editability. This can result in subtle changes in facial structures that are visually perceptible before and after editing. As shown in Fig.~\ref{fig:ab} (b) fourth column, if left unaddressed, this can lead to inconsistent facial structures between the generated images and the source images after makeup transfer, which is not aligned with the task objective. To tackle this challenge, we propose a training strategy that decouples content and structure to eliminate the impact of inconsistent facial Structures between training data pairs and ensure the generated images have consistent facial structures with the source images.

Specifically, during training, $\mathit{I_s}$ and $\mathit{I_c}$ obtained from the target image are fed into the content encoder and structural encoder, respectively, while the augmented $\mathit{I_t}$ is used as the input for the makeup encoder. This implies that the facial structures of the final generated target image are determined by the structural encoder, while the content is determined by the content encoder. The pre-trained U-Net is used to predict and remove the noise that needs to be eliminated for reconstructing the target image. By decoupling content and structure during training, we can control and adjust the content and structure separately, improving the accuracy and consistency of the generated images. During inference,  $\mathit{I_s}$ and $\mathit{I_c}$ obtained from the source image are input to the content encoder and structural encoder respectively.
\vspace{-0.1cm}
\subsubsection{Augmentation.}
In our training pipeline, we have used a variety of augmentations, which are crucial for adapting to real-world scenarios and achieving successful makeup transfer. These augmentations can be divided into two categories: \textbf{makeup augmentations} and \textbf{source augmentations}. Makeup augmentations involve structural augmentations such as face warping and affine transformations applied to the input of the makeup encoder. The purpose of these augmentations is to eliminate Content information from the reference makeup image and to disrupt the pixel alignment between the reference makeup image and the source image. Source augmentations are designed to make our model adaptable to varying facial sizes and poses in the source images in real-world scenarios. These augmentations include synchronized affine transformations applied to both the source image and the target image.

\subsubsection{Loss function.}
Our loss function is similar to the original Stable Diffusion training objective function and can be mathematically represented as:
\begin{equation}
    L(\bm{\theta}) := \mathbb{E}_{\mathbf{x_0}, t, \bm{\epsilon}}\left[\left\|\bm{\epsilon}-\bm{\epsilon_\theta}\left(\mathbf{x_t}, t,\mathbf{c_i},\mathbf{c_e},\mathbf{c_m}\right)\right\|_2^2\right],
\end{equation}
where $\mathbf{x_t}$ is a noisy image latent constructed by adding noise $\bm{\epsilon} \in \mathcal{N}(\mathbf{0},\mathbf{1})$  to the image latents $\mathbf{x_0}$ and the network $\bm{\epsilon_\theta(\cdot)}$ is trained to predict the added noise, $\mathbf{c_i},\mathbf{c_e},\mathbf{c_m}$ represent the content condition input, structural condition input, and makeup condition input, respectively.

\begin{figure*}[!h]
\centering
\includegraphics[width=0.8\linewidth]{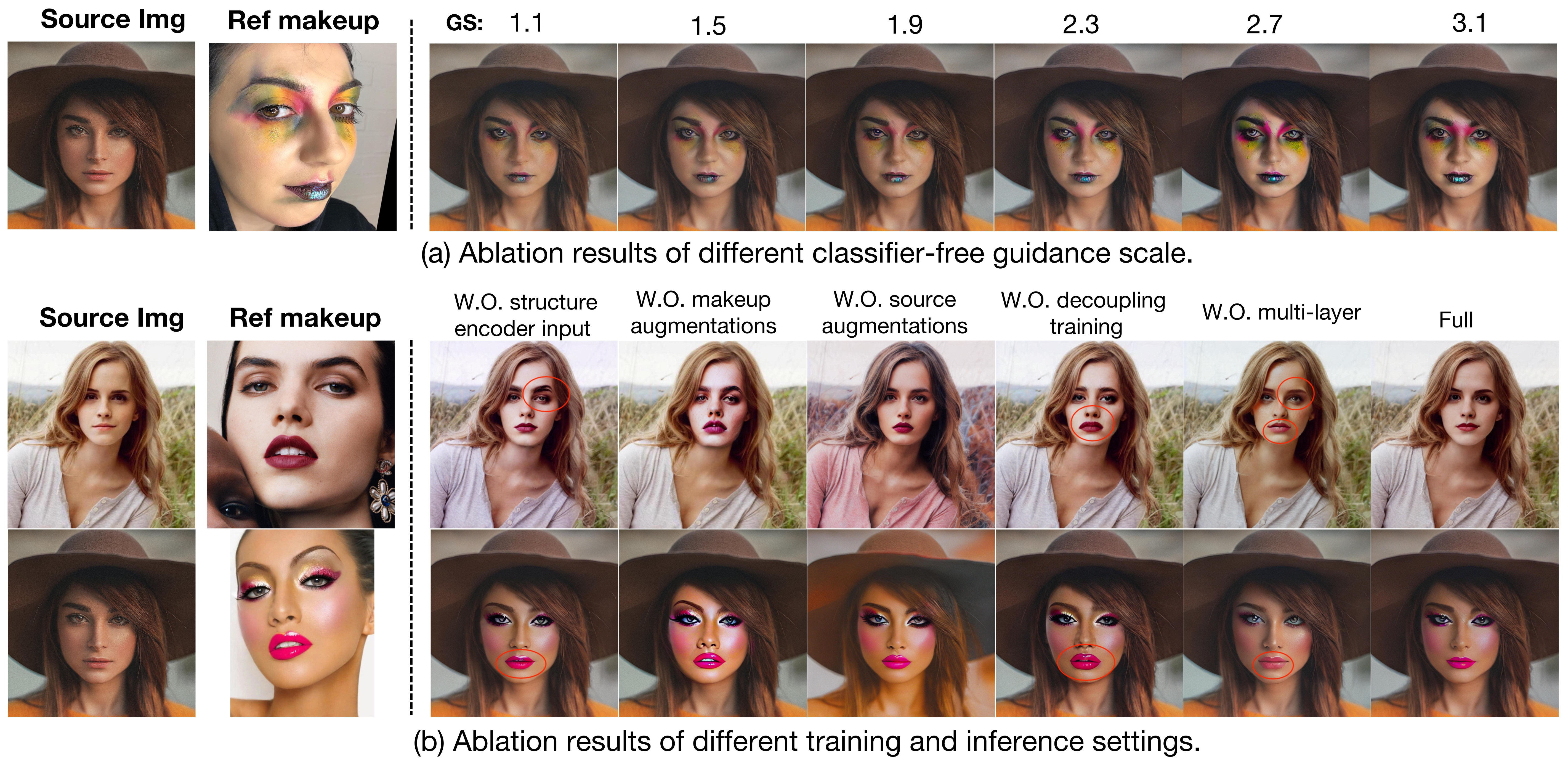}
\vspace{-0.4cm}
  \caption{Qualitative ablation results. Figure (a) explores the classifier-free guidance scale. Figure (b) presents the ablation study of different training and inference settings.
  }
  \vspace{-0.5cm}
    \label{fig:ab}
\end{figure*}

\section{Experiments}
\label{exper}

\setlength{\tabcolsep}{0.4mm}{
\begin{table}[!t]
\begin{scriptsize}
\centering
\begin{tabular}{*{2}c|cccc|cccc}
\toprule
& \multirow{2}*{Method} & CLIP-I $\uparrow$ & DINO-I $\uparrow$ & SSIM  $\uparrow$ & L2-M $\downarrow$ & CLIP-I $\uparrow$ & DINO-I $\uparrow$ & SSIM  $\uparrow$ & L2-M $\downarrow$ \\
& &\multicolumn{4}{c}{(M-wild dataset)} &\multicolumn{4}{c}{(CPM-real dataset)} \\

\midrule
&BeautyGAN &0.742 &0.926 &0.884 &48.408 &0.718 & 0.915 &0.866 &49.168\\
&CPM &0.754 &0.920 &0.873  &94.092  &\textcolor{blue}{0.757} & \textcolor{blue}{0.920}&0.846 &95.172\\
&PSGAN &0.748 &0.928 &0.862 &45.742 &0.724 &0.916 &0.829 &46.091\\
&EleGANt &0.759 &0.928 &0.878 &44.463 &0.731 &0.917 &0.838 &45.797\\
&SCGAN &0.751 &0.928 &0.882 &44.296 &0.726 &0.916 &0.858 &46.284\\
&SPMT &\textcolor{blue}{0.761} &\textcolor{blue}{0.931} &\textcolor{red}{0.915} &\textcolor{blue}{29.126} &0.735 &0.920 &\textcolor{red}{0.900} &\textcolor{blue}{28.299}\\
&SSAT &0.758 &0.930 &\textcolor{blue}{0.914} &32.578 &0.719 & 0.915&\textcolor{blue}{0.896} &33.604\\
&Ours &\textcolor{red}{0.768} &\textcolor{red}{0.933} &0.910  &\textcolor{red}{29.087} &\textcolor{red}{0.760} &\textcolor{red}{0.923} &0.868 &\textcolor{red}{27.976}\\
\bottomrule
\end{tabular}
\end{scriptsize}
\caption{\textbf{Quantitative comparison} of different methods. Metrics that are red and blue represent methods that rank 1st and 2nd, respectively.}
\label{tab}
\vspace{-0.7cm}
\end{table}
}

\subsection{Experimental Setup}
We employed Stable Diffusion V1-5 as the pre-trained diffusion model and trained it on the Makeup-real Dataset, which consists of 20k makeup-non-makeup pairs, and each pair with the same identity. We focus on real-world conditions and utilize extremely aggressive augmentation. While training, the parameters of the D-P makeup encoder, makeup cross-attention layers, content encoder, and structural encoder are updated, while the pre-trained text-to-image model remains frozen. The model underwent 100,000 iterations of training on 8 H800 GPUs, with a batch size of 16 per GPU and a learning rate of 5e-5. Inference was performed using DDIM as the sampler, with a step size of 30 and a guidance scale set to 1.5.

\subsection{Evaluation Metrics}
To provide a thorough and objective assessment of the performance of each algorithm in different aspects of makeup transfer, we employed the following metrics: \textbf{CLIP-I}\cite{clip} and \textbf{DINO-I}\cite{dinov1} were utilized to assess makeup transfer performance. After makeup transfer, the content and structure information of the input image should not be changed. So we use \textbf{SSIM}~\cite{ssim} and \textbf{L2-M} to evaluate the degree of content and structural maintenance.

Notably, \textbf{CLIP-I} and \textbf{DINO-I} are metrics used to calculate the cosine similarity between the image embeddings of the target image and the makeup image. They utilize the CLIP image encoder or DINO backbone to extract image features. \textbf{L2-M} metric utilizes the facial segmentation method to obtain the background images of the target and the source image, calculate the square of the difference, and then average it based on the number of pixels.

\subsection{Comparison Methods}
We performed comprehensive comparisons with the most representative makeup transfer algorithms currently available, including
BeautyGAN\cite{li2018beautygan},
CPM\cite{nguyen2021lipstick},
PSGAN\cite{jiang2020psgan}, 
EleGANt\cite{yang2022elegant},
SCGAN\cite{deng2021spatially}, SPMT\cite{Zhu2022SemiparametricMT} and 
SSAT\cite{sun2023ssat}, and conduct comparisons on M-Wild dataset\cite{jiang2020psgan} and CPM-real dataset\cite{nguyen2021lipstick}  respectively. 

\subsection{Experiment Results}
\label{exp}
\subsubsection{Qualitative Comparison.}

\begin{figure*}[!t]
\centering
\includegraphics[width=0.85\linewidth]{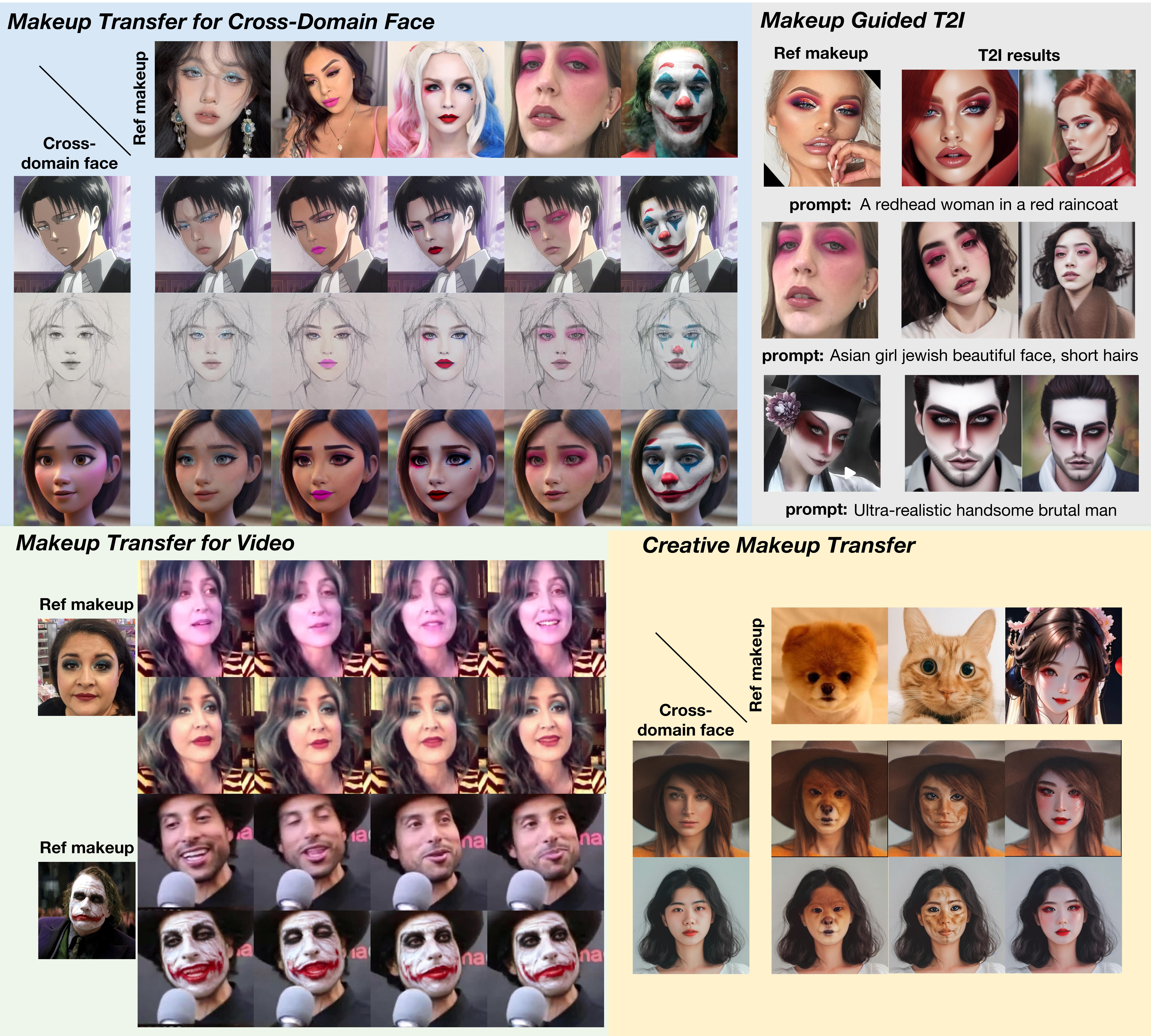}
\vspace{-0.3cm}
  \caption{The visual results of Stable-Makeup demonstrate its robustness and generalizability for various applications.
}
\vspace{-0.4cm}
    \label{fig:app}
\end{figure*}

Fig.~\ref{fig:comparison} provides a visual comparison of our proposed Stable-Makeup method and other makeup transfer methods. To further evaluate the performance of these methods on various makeup styles, we used some popular makeup images from social media as reference makeup in addition to the test set. Overall, most methods maintained good consistency in content and structure, with only CPM introducing some artifacts during makeup transfer. In terms of makeup transfer capability, our method outperformed other methods by a significant margin. Other methods were only effective in transferring simple lip and eye shadow colors and were only effective for light makeup. In contrast, our method not only performed well for light makeup but also effectively aligned and transferred heavy makeup to the corresponding positions in the source image, while preserving makeup details and corresponding color distributions on the facial regions of the makeup image.

\subsubsection{Quantitative Comparison.}

To test the ability of methods in the real-world environment, We utilized all makeup images from M-wild Dataset and CPM-Real dataset as reference images, while randomly selecting non-makeup images from the MT-dataset and MWild-Dataset as source images. We utilized a total of 7017 makeup reference images, paired with an equal number of source images.

As shown in Table~\ref{tab}, It is evident that Stable-Makeup outperforms all other makeup transfer methods in terms of the DINO-I and CLIP-I metrics, indicating that our method has superior makeup transfer capabilities compared to other methods. Additionally, the SSIM and L2-M metrics achieved good scores when compared with other methods, indicating that our method is able to maintain good content and structure information in the transferred images. It should be noted that SSIM metrics may not accurately reflect the makeup transfer performance, as they may produce good scores when the transferred result is identical to the source image, indicating that the transfer algorithm did not work. In line with previous studies\cite{nguyen2021lipstick, xiang2022ramgan, Zhu2022SemiparametricMT}, we also listed it as a reference in our evaluation.

\subsubsection{Ablation Study.}
Firstly, we explored the effect of different parameters for the classifier-free guidance scale (GS) under makeup conditions in figure~\ref{fig:ab} (a). Next, in figure~\ref{fig:ab} (b), we conducted ablation experiments by systematically removing each module using the same experimental setup. As shown in figure~\ref{fig:ab} (a), it was found that a larger GS led to a greater degree of makeup transfer, while our method maintained good content and structural consistency of the source image under different guidance parameters. In figure~\ref{fig:ab} (b), it demonstrated that without the structural encoder input during inference (first column), the inconsistency in facial structure was observed, such as changes in the shape of eyebrows and mouth in the first column of the generated images. Without makeup augmentation during training (second column), the model does not possess the capability to transfer makeup, but rather, it solely transfers the facial region of the makeup image to the target image. Removing source augmentation during training (third column), the alignment and transferability of the model were weakened, and the background color was changed. When removing the structural encoder during training (fourth column, W.O. decoupling training), the facial structure also could not be well preserved. Without the multi-layer strategy for the makeup encoder (fifth column), the transfer of makeup details was incomplete. When our method was fully implemented, it achieved a complete transfer of makeup details while maintaining good consistency of the content and structure of the source image. Furthermore, we provide additional quantitative ablation results in
supplementary.

\subsection{More Applications}
As illustrated in Fig.~\ref{fig:app}, our Stable-Makeup not only achieves unprecedented makeup transfer effects on real faces but also enables various applications that were previously unattainable with traditional makeup transfer methods. For instance, our method can perform makeup transfer on cross-domain faces and guide text-to-image generation with reference makeup conditions to create creative artistic images. Our method can also perform makeup transfer on videos by utilizing effective multi-frame concatenation. Moreover, our approach supports other domain's reference makeup, such as animal-inspired subjects, animated characters and so on. providing a broader range of creative makeup options for real human faces.

\section{Conclusion}
In this paper, we propose the Stable-Makeup, a novel diffusion-based framework aiming for real-world makeup transfer. This method has made a significant breakthrough in the field of makeup transfer by achieving effects that were previously unattainable. The Stable-Makeup consists of three core modules. The detail-preserving makeup encoder captures intricate makeup details, integrating with two modules for content and structural control of the source image. Makeup cross-attention layers align facial regions between reference and source images, while content and structure decoupling training enhances consistency. Additionally, we propose an automated pipeline for generating diverse makeup pairing data for training. Our proposed method has surpassed all existing makeup transfer algorithms and has expanded the boundaries of the makeup transfer field. This breakthrough has significant implications for the practical application of makeup transfer technology in various fields.



\bibliographystyle{IEEEtran}

\clearpage
\appendix
\begin{center} 
\Large \textbf{Supplementary Materials}
\end{center}

\renewcommand\thesection{\Alph{section}}

\section{Preliminaries}
\label{prel}
\subsubsection{Diffusion models.}
Diffusion Model (DM)~\cite{ho2020denoising, scoremodel} belongs to the category of generative models that denoise from a Gaussian prior $\mathbf{x_T}$ to target data distribution $\mathbf{x_0}$ by means of an iterative denoising procedure. Latent Diffusion Model (LDM)~\cite{rombach2022high} is proposed to model image representations in the autoencoder’s latent space. LDM significantly speeds up the sampling process and facilitates text-to-image generation by incorporating additional text conditions. The LDM loss is:
\begin{equation}
    L_{LDM}(\bm{\theta}) := \mathbb{E}_{\mathbf{x_0}, t, \bm{\epsilon}}\left[\left\|\bm{\epsilon}-\bm{\epsilon_\theta}\left(\mathbf{x_t}, t, \bm{\tau_{\theta}}(\mathbf{c})\right)\right\|_2^2\right],
\end{equation}
where $\mathbf{x_t}$ is an noisy image latent constructed by adding noise $\bm{\epsilon} \in \mathcal{N}(\mathbf{0},\mathbf{1})$  to the image latents $\mathbf{x_0}$ and the network $\bm{\epsilon_\theta(\cdot)}$ is trained to predict the added noise, $\bm{\tau_\theta(\cdot)}$ refers to the BERT text encoder~\cite{bert} used to encodes text description $\mathbf{c_t}$.

Stable Diffusion (SD) is a widely adopted text-to-image diffusion model based on LDM. Compared to LDM, SD is trained on a large LAION~\cite{laion5b} dataset and replaces BERT with the pre-trained CLIP~\cite{clip} text encoder.
\subsubsection{ControlNet.} ControlNet~\cite{zhang2023adding} is one of the most popular control modules in current diffusion models. It receives inputs from various modalities as spatial control conditions, guiding the diffusion model to generate desired images, thereby achieving controllable generation. ControlNet copies an identical U-Net structure as trainable parameters and locks the original U-Net parameters, The trainable copy is connected to the locked model with zero convolution layers, denoted $\mathcal{Z}(\cdot ; \cdot)$. The complete ControlNet can represented as:
\begin{equation}
    \boldsymbol{y}_{\mathrm{c}}=\mathcal{F}(\boldsymbol{x} ; \Theta)+\mathcal{Z}\left(\mathcal{F}\left(\boldsymbol{x}+\mathcal{Z}\left(\boldsymbol{c} ; \Theta_{\mathrm{z} 1}\right) ; \Theta_{\mathrm{c}}\right) ; \Theta_{\mathrm{z} 2}\right)
\end{equation}

Where $\Theta_{\mathrm{z}1}$ and $\Theta_{\mathrm{z}2}$ represent two different zero conv layers' parameters respectively. $\mathcal{F}$ represent diffusion U-Net, $\boldsymbol{x}$ represent image latent, $\Theta$ is the frozen weight of the U-Net and $\Theta_{\mathrm{c}}$ is the trainable copy weight of the U-Net.

\subsubsection{CLIP.}
CLIP~\cite{clip} consists of two integral components: an image encoder, represented as $F(x)$, and a text encoder, represented as $G(t)$. The image encoder, $F(x)$, transforms an image $x$ with dimensions $\mathbb{R}^{3 \times H \times W}$ (height $H$ and width $W$) into a $d$-dimensional image feature $f_x$ with dimensions $\mathbb{R}^{N \times d}$, where $N$ is the number of divided patches. On the other hand, the text encoder, $G(t)$, creates a $d$-dimensional text representation gt with dimensions $\mathbb{R}^{M \times d}$ from natural language text $t$, where $M$ is the number of text prompts. After training on a contrastive loss function, CLIP can be applied directly for zero-shot image recognition without the need for fine-tuning the entire model.

\section{Evaluation Benchmarks}
\begin{itemize}
    \item \textbf{CPM-real dataset}: It is very diverse in terms of makeup styles, containing both color and pattern makeup. The degree of makeup can vary from light to heavy, from color-oriented to pattern-driven.
    \item \textbf{M-wild dataset:} 
    It contains facial images with various poses and expressions as well as complex backgrounds to test methods in a real-world environment.
\end{itemize}

\setlength{\tabcolsep}{0.5mm}{
\begin{table}[t]
\begin{small}
\centering
\begin{tabular}{*{2}c|cccc}
\toprule
& \multirow{2}*{Ablation Setups} & CLIP-I $\uparrow$ & DINO-I $\uparrow$ & SSIM  $\uparrow$ & L2-M $\downarrow$ \\
& &\multicolumn{4}{c}{(CPM-real dataset)} \\
\midrule
&W.O. structure input& 0.716 &0.913 &0.759 &97.746 \\
&W.O. source aug &0.752 &0.911 &0.858 &96.385 \\
&W.O decoupling training &0.745 &0.917 &0.754 &35.675 \\
&W.O multi-layer &0.742 &0.907 &0.861 &29.076 \\
&Ours full &\textbf{0.760} &\textbf{0.923} &\textbf{0.868}  &\textbf{27.976}\\
\bottomrule
\end{tabular}
\captionof{table}{Quantitative ablation results on CPM-real dataset.}
\label{abla}
\end{small}
\end{table}
}

\section{Quantitative Ablation}
Quantitative ablation results are shown in Table~\ref{abla}. We performed quantitative ablation experiments using the same experimental setup on the CPM-real dataset as the experiments section. We observed that our method achieved optimal performance when all modules were retained while removing the structural encoder input during inference (first row) or training (third row) led to a significant decrease in content and structure consistency. Removing the multi-layer strategy during training resulted in a significant decrease in makeup transfer ability, and the absence of source enhancement led to a significant decrease in background consistency. We exclude the setting without makeup augmentations, as shown in Figure 5 (b) second column in our paper, where the model can only perform face swapping and lacks the ability to perform makeup transfer. This could significantly affect the calculation of our makeup transfer metrics.

\section{Further Exploration of Data Construction}
\label{sup:data}
\begin{figure}[t]
    \centering
    \includegraphics[width=1\linewidth]{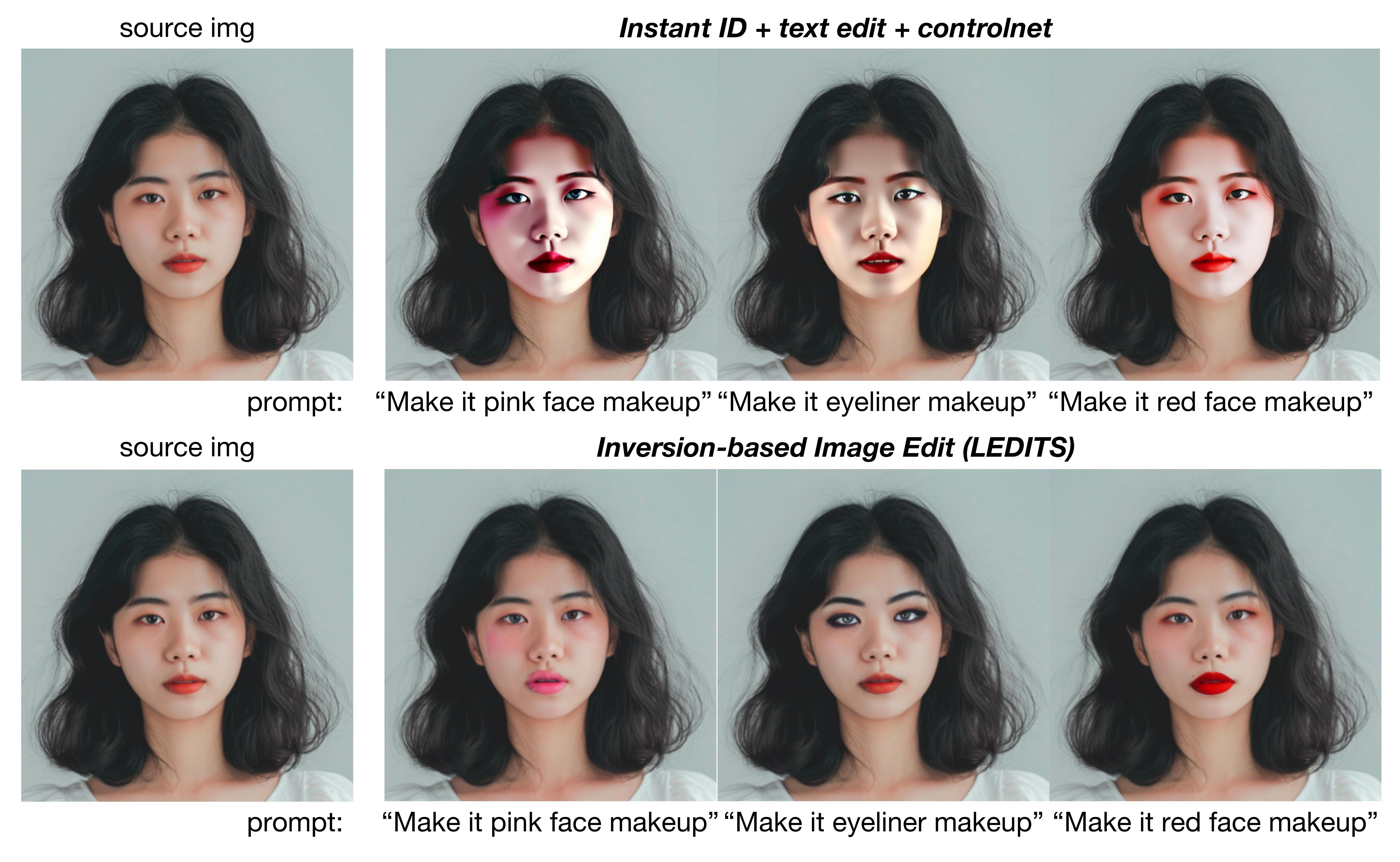}
    \caption{Visual results of different data construct pipelines.}
    \label{fig:makeup}
\end{figure}

Currently, diffusion-based methods have demonstrated remarkable performance in terms of fidelity and diversity, making them a crucial tool for achieving high-quality and diverse data in step 2 of our data construction process. In addition to the inversion-based image editing methods we employed, we could also choose to use existing identity customization methods, such as InstantID\cite{wang2024instantid}, which do not require test-time fine-tuning. However, as shown in figure~\ref{fig:makeup}, when using these diffusion-based methods for editing facial images, visible changes in facial structure may occur during makeup generation guided by text. For example, the mouth shape of these generated images cannot be exactly the same. Given the current limitations of diffusion-based methods in terms of fine-grained control, our proposed content and structure decoupling strategy is particularly important for maintaining facial expression and structure.

\section{Designing Choice of Detail-Preserving Makeup Encoder}
\label{sup:encoder}
In this section, we conducted a comparative study between different mapping structure settings under the same training conditions to demonstrate that mapping without preserving spatial information struggles to align and preserve makeup details at a fine-grained level, and to validate the effectiveness of our proposed method. We first introduced our experimental settings in Section~\ref{sup:setup}. Subsequently, we conduct a visual comparison in Sec.~\ref{sup:results} to demonstrate the pivotal role of our proposed method in preserving and aligning the makeup details. Additionally, in Sec.~\ref{sup:attn} we further demonstrated the effectiveness of our proposed method in preserving and aligning the makeup details by visualizing the attention maps of the makeup cross-attention layers in U-Net.

\subsection{Experimental Setup}

\begin{figure*}[!h]
    \centering
    \includegraphics[width=1\linewidth]{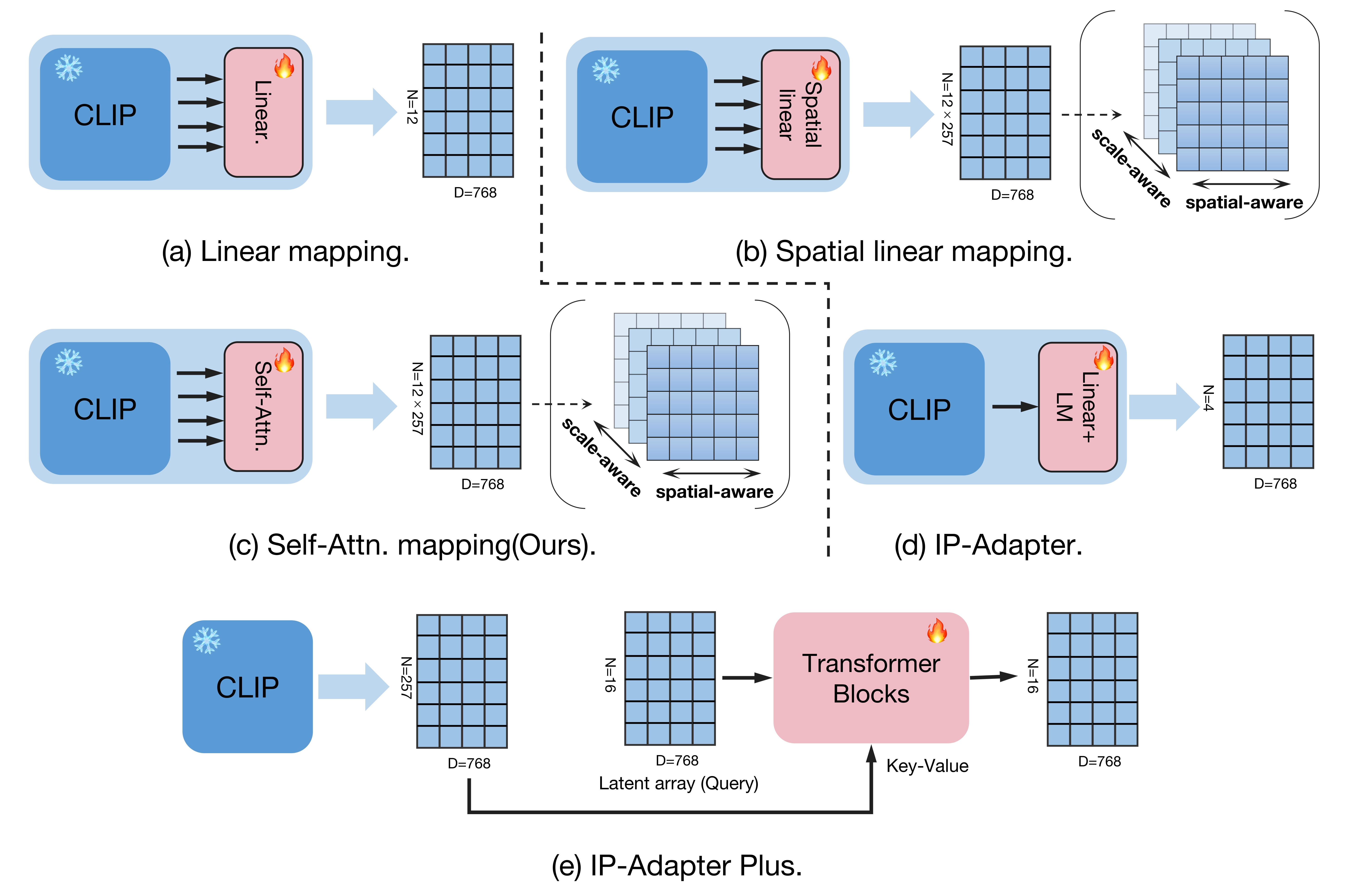}
    \caption{The structural composition of the five distinct mapping schemes. We omitted the drawing of the global vector (cls token from CLIP) for the sake of brevity and clarity.}
    \label{fig:sup_des}
\end{figure*}

\label{sup:setup}
As shown in Fig.~\ref{fig:sup_des}, we conducted training on five sets of mapping schemes. 
\begin{itemize}
    \item Linear mapping: We utilized linear layers to project the spatial dimension to a smaller dimension of (1, 12, 1024), where the original dimension was (1, 257$\times$12, 1024).
    \item Spatial linear mapping: We utilized linear layers to project the feature dimension to a smaller dimension without affecting the spatial features dimension, where the original dimension was (1, 257$\times$12, 1024), and the projected dimension was (1, 257$\times$12, 768).
    \item Self-Attention mapping: We utilized our proposed self-attention mapping in the Methodology section. The projected dimension is the same as the original dimension (1, 257$\times$12, 1024).
    \item IP-Adapter: We replaced our D-P makeup encoder with an IP-Adapter that projects the pooled feature of CLIP (1, 1, 768) to a higher dimension features (1, 4, 768) using linear layer and layer norm.
    \item IP-Adapter Plus: We replaced our D-P makeup encoder with an IP-Adapter Plus, which use a small query network to learn features. Specifically, 16 learnable tokens are defined to extract information from the grid features using a lightweight transformer model. The token features from the query network serve as input to the cross-attention layers. Where the original dimension was (1, 257, 1024), and the final dimension was (1, 16, 768).
\end{itemize}
All of these models were trained on the Makeup-real dataset using OpenCLIP ViT-L/14 as the CLIP model and StableDiffusion V1-5 as the pre-trained diffusion model. The training parameters were kept consistent, with a learning rate of 5e-5, batch size of 8, and training on 2 H800s. It is noteworthy that only spatial linear mapping and our self-attention mapping preserve the spatial information of the makeup image features among the five mapping settings, while the others introduce varying degrees of spatial distortion to the makeup image features.
\begin{figure}[!h]
    \centering
\includegraphics[width=1\linewidth]{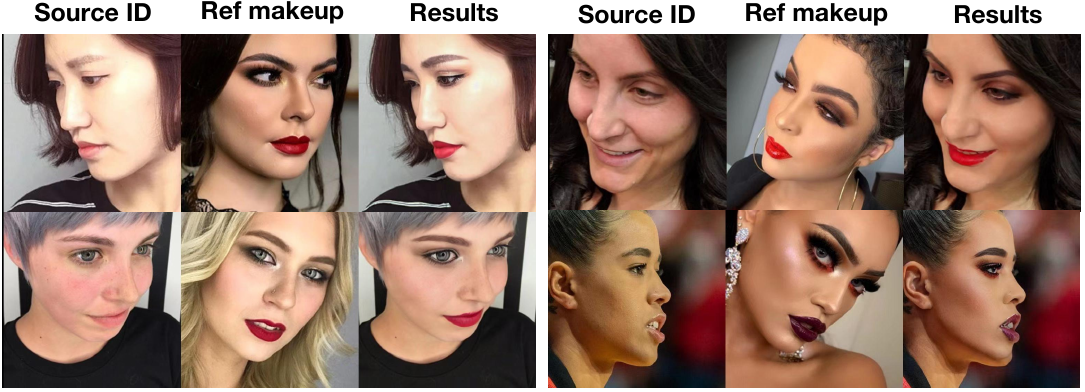}
  \caption{Results under extreme viewing angles. (please zoom in to better view).}
    \label{fig:supexm}
\end{figure}

\begin{figure*}[!h]
    \centering
    \includegraphics[width=0.9\linewidth]{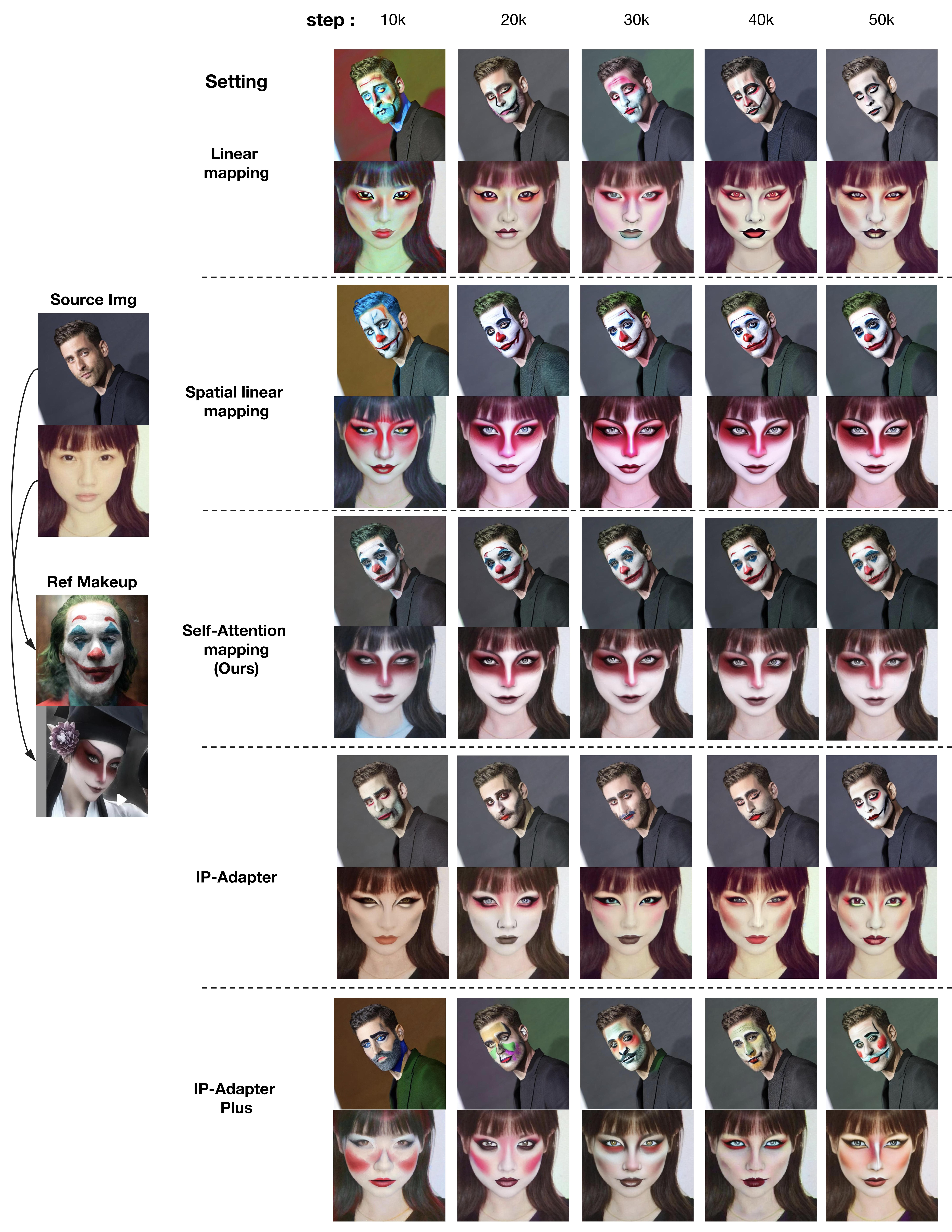}
    \caption{Visual comparison results under the same training step but with different mapping settings.}
    \label{fig:sup_com}
\end{figure*}

\subsection{Comparison}
\label{sup:results}
Under the same training conditions, the makeup transfer results at different training steps are presented in figure~\ref{fig:sup_com}. Overall, our self-attention mapping achieves the most precise makeup transfer and the smallest number of model training convergence steps. Comparing the results under different settings, it is observed that spatial linear mapping, IP-Adapter, and IP-Adapter plus all maintain the content and structure consistency of the source image, but they fail to align and transfer the details of the reference makeup image to the facial region of source image. Both spatial linear mapping and our self-attention mapping can align and transfer the makeup, but our method outperforms spatial linear mapping in terms of the precision of makeup transfer. For instance, our method completely preserves the red graffiti on the forehead of the clown (in the fifth line), while spatial linear mapping fails to do so (in the third line). It is noteworthy that our method achieves good transfer performance at the 20,000th training step, while spatial linear mapping struggles to preserve the makeup details. These experimental results effectively demonstrate the importance of preserving spatial information for preserving fine makeup details.

\subsection{Visualization of Attention maps }
\label{sup:attn}

\begin{figure*}[t]
    \centering
    \includegraphics[width=1\linewidth]{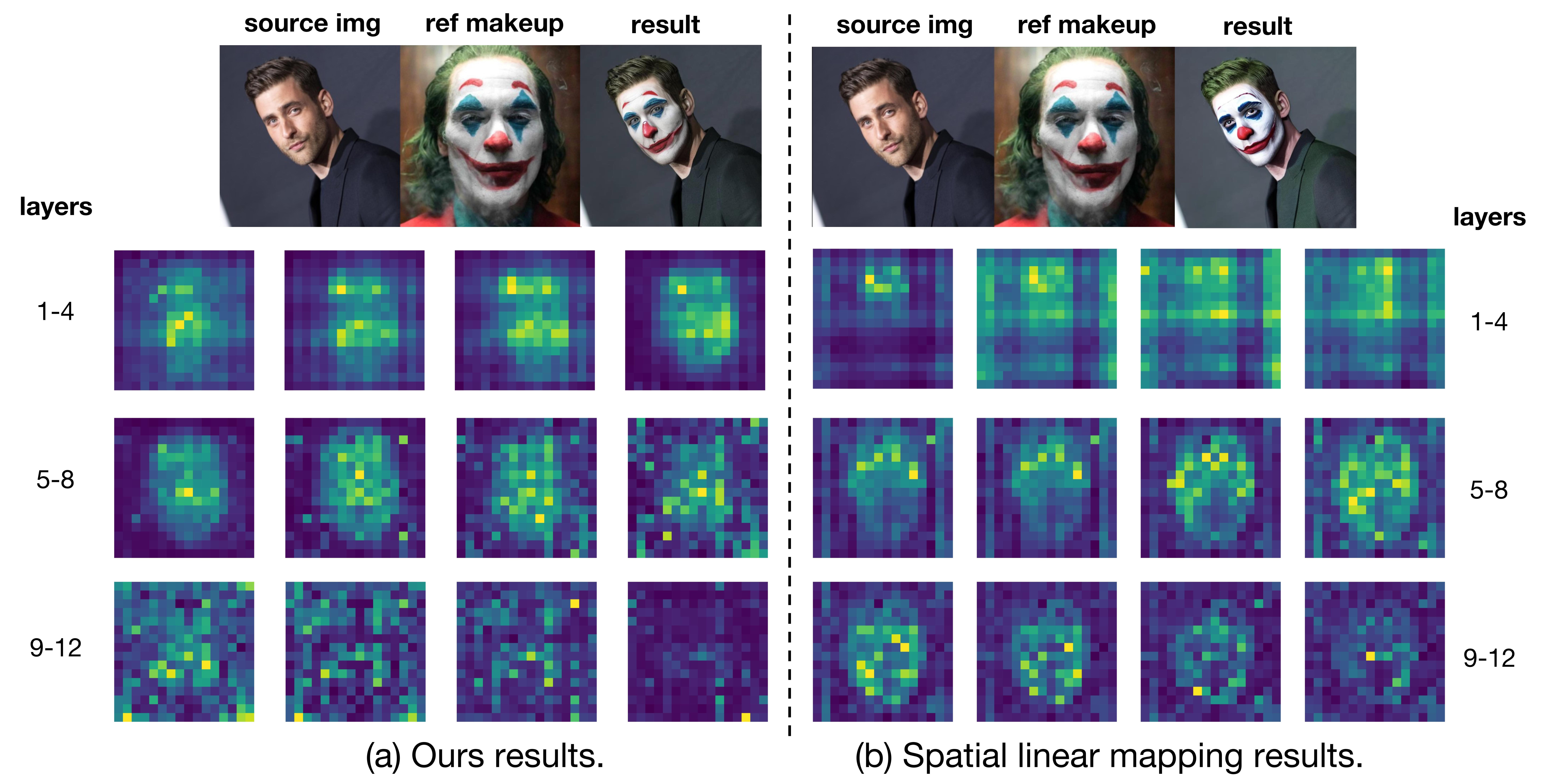}
    \caption{Visualization of attention maps with different mapping settings at the 50,000th training step.}
    \label{fig:sup_attn}
\end{figure*}

In order to conduct a more comprehensive comparison between spatial linear mapping and self-attention mapping, we visualized the attention maps of the makeup cross-attention layers under both configurations at the 50,000th training step in this experiment. Specifically, we extract and aggregate the attention matrices calculated from the patch features of the source facial region (Query) and the multi-scale makeup embeddings (Key-Value), based on the index of the facial position, and visualize them at different layers, as shown in Figure~\ref{fig:sup_attn} (a). From the attention maps, we can see that the self-attention layer can effectively weigh the features, resulting in a high response for the facial patch features in between the source image and the makeup image, while a low response between the other positions in the makeup image. However, when we replace the self-attention mapping with spatial linear mapping, the facial region of the source image also exhibits a high response to the background features, as shown in the attention maps of layers 1-4 in Figure~\ref{fig:sup_attn} (b). These findings highlight the advantages of self-attention mapping in achieving more precise and effective feature weighting, which is crucial for achieving high-quality makeup transfer results.

\section{More Results}
\begin{figure*}[t]
    \centering
    \includegraphics[width=0.9\linewidth]{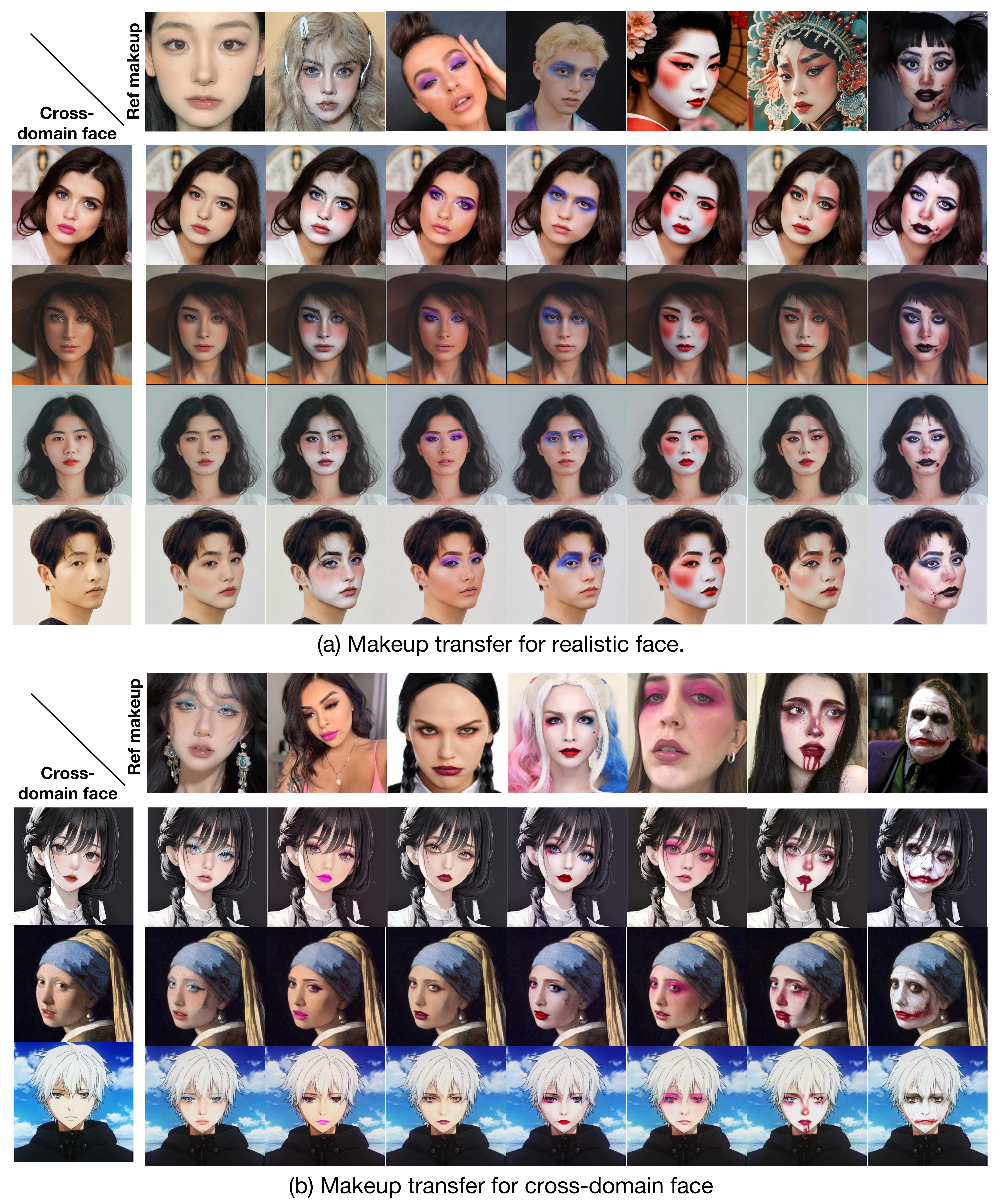}
    \caption{More results of makeup transfer for realistic faces and cross-domain faces.}
    \label{fig:sup_mt}
\end{figure*}

In this section, we present a range of additional results that demonstrate the robustness and superiority of our approach. Specifically, as shown in Fig.~\ref{fig:sup_mt} and Fig.~\ref{fig:sup_vi}, we showcase the effectiveness of our method in makeup transfer on realistic faces, makeup transfer on cross-domain faces, video makeup transfer, and makeup-guided text-to-image generation. These results provide compelling evidence of the effectiveness of our approach and highlight its potential for a wide range of applications.

\section{User study}
\setlength{\tabcolsep}{0.5mm}{
\begin{table}[t]
\begin{small}
\centering
\caption{\textbf{User study.} Metrics that are bold represent methods that rank 1st. Our method achieves the highest average score.}
\label{user}
\begin{tabular}{*{2}c|cccccccc}
\toprule
& \scriptsize{Method} & \scriptsize{BeautyGAN} & \scriptsize{CPM} & \scriptsize{PSGAN} & \scriptsize{EleGANt} & \scriptsize{SCGAN} & \scriptsize{SPMT} & \scriptsize{SSAT} & \scriptsize{Ours}  \\
\midrule
&\scriptsize{Average Score} &3.01 &3.13 &2.95 &3.62 &2.90 &3.35 &2.85 &\textbf{4.58}
\\
\bottomrule
\end{tabular}
\end{small}
\end{table}
}
We conducted a user study to compare our method with other methods perceptually.  We randomly select 10 source images and 10 reference images from both the M-Wild dataset and the CPM-real dataset. For each evaluation, each user will see one source image with a reference makeup image and one image generated by each method. 60 evaluators were asked to provide a comprehensive score for each generated image based on its makeup consistency, content and structure consistency, using a scale of 1 (worst) to 5 (best). The results of average scores in Table~\ref{user} indicate that our method outperforms the comparison methods. Our method achieves the highest average score, which clearly indicates that our method outperforms the comparison methods.
\section{Extreme viewing angles}
Our approach is capable of processing source images under extreme viewing angles (in Fig.~\ref{fig:supexm}).

\section{Broader Impact}
The broader impact of diffusion-based makeup transfer methods is significant, as they have the potential to revolutionize the beauty industry by enabling more efficient and personalized makeup applications. However, it is important to consider the ethical implications of such technology, including issues related to privacy, consent, and the potential perpetuation of beauty standards. We do not condone the unauthorized use of our method to edit the makeup of others' photos. As with any emerging technology, we urge caution in the use of diffusion-based makeup transfer methods and further consideration of their ethical and legal implications.

\section{Limitations}
Due to potential inconsistencies in facial structure between paired data in the training dataset, which may arise from conflicts between text editability and maintaining the source image structure in current text-based editing methods, this issue may potentially impact the performance of our model. To address this, we plan to refine the data selection process and manually customize the data. In the future, we also plan to extend our method to 3D tasks.

\begin{figure*}[t]
    \centering
    \includegraphics[width=0.9\linewidth]{images/sup_video.jpg}
    \caption{More results of video makeup transfer and makeup guided text-to-image generation.}
    \label{fig:sup_vi}
\end{figure*}

\end{document}